\documentclass[sigconf]{acmart}

\usepackage{booktabs} 
\usepackage{times}
\usepackage{latexsym}
\usepackage{mathtools}
\usepackage{amsfonts}
\usepackage{graphicx}
\usepackage{nicefrac}
\usepackage{url}
\usepackage{commath}
\usepackage{array}
\usepackage{anyfontsize}
\usepackage{tabularx}
\usepackage{amsmath,enumitem}

\newcolumntype{x}[1]{>{\centering\arraybackslash}p{#1}}
\newcolumntype{Y}{>{\centering\arraybackslash}X}

\copyrightyear{2019}
\acmYear{2019} 
\setcopyright{iw3c2w3}
\acmConference[WWW '19]{Proceedings of the 2019 World Wide Web Conference}{May 13--17, 2019}{San Francisco, CA, USA}
\acmBooktitle{Proceedings of the 2019 World Wide Web Conference (WWW '19), May 13--17, 2019, San Francisco, CA, USA}
\acmPrice{}
\acmDOI{10.1145/3308558.3313395}
\acmISBN{978-1-4503-6674-8/19/05}

\usepackage{balance}

\begin{document}

\title{Learning Graph Pooling and Hybrid Convolutional Operations for Text Representations}

\author{Hongyang Gao}
\affiliation{%
  \institution{Texas A\&M University}
  \city{College Station}
  \state{TX}
  \postcode{77843}
}
\email{hongyang.gao@tamu.edu}

\author{Yongjun Chen}
\affiliation{%
  \institution{Washington State University}
  \city{Pullman}
  \state{WA}
  \postcode{99164}
}
\email{yongjun.chen@wsu.edu}

\author{Shuiwang Ji}
\affiliation{%
  \institution{Texas A\&M University}
  \city{College Station}
  \state{TX}
  \postcode{77843}
}
\email{sji@tamu.edu}

\renewcommand{\shortauthors}{H. Gao et al.}

\begin{abstract}
With the development of graph convolutional networks (GCN), deep
learning methods have started to be used on graph data. In
additional to convolutional layers, pooling layers are another
important components of deep learning. However, no effective pooling
methods have been developed for graphs currently. In this work, we
propose the graph pooling~(gPool) layer, which employs a trainable
projection vector to measure the importance of nodes in graphs. By
selecting the $k$-most important nodes to form the new graph, gPool
achieves the same objective as regular max pooling layers operation
on images and texts. Another limitation of GCN when used on graph-based text
representation tasks is that, GCNs do not consider the order
information of nodes in graph. To address this limitation, we
propose the hybrid convolutional~(hConv) layer that combines GCN and
regular convolutional operations. The hConv layer is capable of
increasing receptive fields quickly and computing features
automatically. Based on the proposed gPool and hConv layers, we
develop new deep networks for text categorization tasks. Our
experimental results
show that the networks based on gPool and hConv layers achieves new
state-of-the-art performance as compared to baseline methods.
\end{abstract}

%
%



\newcommand*{\origrightarrow}{}
\let\oldarrow\textrightarrow
\renewcommand*{\textrightarrow}{\fontfamily{cmr}\selectfont\origrightarrow}

\maketitle

\section{Introduction}

Convolutional neural networks~(CNNs)~\cite{lecun1998gradient} have
shown great capability of solving challenging tasks in various
fields such as computer vision and natural language
processing~(NLP). A variety of CNN networks have been proposed to
continuously set new performance
records~\cite{krizhevsky2012imagenet,simonyan2015very,he2015deep,huang2017densely}.
In addition to image-related tasks, CNNs are also successfully
applied to NLP tasks such as text
classification~\cite{zhang2015character} and neural machine
translation~\cite{bahdanau2014neural,vaswani2017attention}. The power of CNNs lies in
trainable local filters for automatic feature extraction. The
networks can decide which kind of features to extract with the help
of these trainable local filters, thereby avoiding hand-crafted
feature extractors~\cite{wang2012end}.

One common characteristic behind the above tasks is that both images
and texts can be represented in grid-like structures, thereby
enabling the application of convolutional operations. However, some
data in real-world can be naturally represented as graphs such as
social networks. There are primarily two types of tasks on graph
data; those are, node
classification~\cite{kipf2016semi,velivckovic2017graph} and graph
classification~\cite{niepert2016learning,wu2017multiple}. Graph
convolutional networks~(GCNs)~\cite{kipf2016semi} and graph
attention networks~(GATs)~\cite{velivckovic2017graph} have been
proposed for node classification tasks under both transductive and
inductive learning settings. However, GCNs lack the capability of
automatic high-level feature extraction, since no trainable local
filters are used.

In addition to convolutional layers, pooling layers are another
important components in CNNs by helping to enlarge receptive fields
and reduce the risk of over-fitting. However, there are still no
effective pooling operations that can operate on graph data and
extract sub-graphs. Meanwhile, GCNs have been applied on text data
by converting texts into
graphs~\cite{schlichtkrull2017modeling,liu2018matching}. Although
GCNs can operate on graphs converted from texts, they ignore the
ordering information between nodes, which correspond to words in
texts.

In this work, we propose a novel pooling layer, known as the graph
pooling~(gPool) layer, that acts on graph data. Our method employs a
trainable projection vector to measure the importance of nodes in a
graph. Based on measurement scores, we rank and select $k$-largest
nodes to form a new sub-graph, thereby achieving pooling operation
on graph data. When working on graphs converted from text data, the
words are treated as nodes in the graphs. By maintaining the order
information in nodes' feature matrices, we can apply convolutional
operations to feature matrices. Based on this observation, we
develop a new graph convolutional layer, known as the hybrid
convolutional~(hConv) layer. Based on gPool and hConv layers, we
develop a shallow but effective architecture for text modeling
tasks~\cite{long2015fully}. Results on text classification tasks
demonstrate the effectiveness of our proposed methods as compared to
previous CNN models.


\section{Related Work}

Before applying graph-based methods on text data, we need to convert
texts to graphs. In this section, we discuss related literatures on
converting texts to graphs and use of GCNs on text data.

\begin{figure}[t]
\centering\includegraphics[width=0.8\columnwidth]{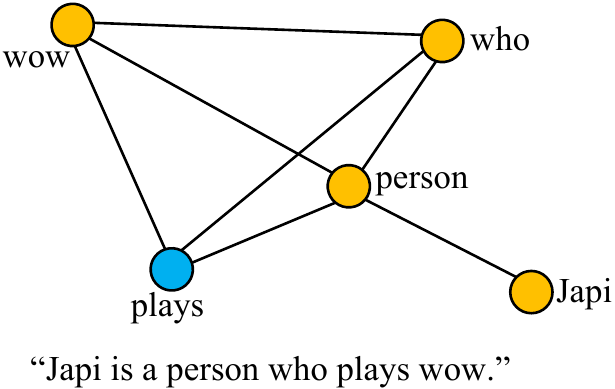}
\caption{Example of converting text to a graph using the
graph-of-words method. For this text, we use noun, adjective, and
verb as terms for node selection. The words of ``Japi", ``person",
``who", ``plays", and ``wow"  are selected as nodes in the graph. We
employ a sliding window size of 4 for edge building. For instance,
there is an undirected edge between ``Japi'' and ``person'', since
they can be covered in the same sliding window in the original
text.} \label{fig:method}
\end{figure}

\subsection{Text to Graph Conversion}

Many graph representations of texts have been explored to capture
the inherent topology and dependence information between words.
\citet{bronselaer2013approach} employed a rule-based classifier to
map each tag onto graph nodes and edges. The tags are acquired by
the part-of-speech (POS) tagging techniques.
In~\cite{liu2018matching} a concept interaction graph representation
is proposed for capturing complex interactions among sentences and
concepts in documents. The graph-of-word
representation~(GoW)~\cite{mihalcea2004textrank} attempts to capture
co-occurrence relationships between words known as terms. It was
initially applied to text ranking task and has been widely used in
many NLP tasks such as information
retrieval~\cite{rousseau2013graph}, text
classification~\cite{rousseau2015text,malliaros2015graph}, keyword
extraction~\cite{rousseau2015main,tixier2016graph} and document
similarity measurement~\cite{nikolentzos2017shortest}.

Before applying graph-based text modeling methods, we need to
convert texts to graphs. In this work, we employ the
graph-of-words~\cite{mihalcea2004textrank} method for its
effectiveness and simplicity. The conversion starts with the phase
preprocessing such as tokenization and text cleaning. After
preprocessing, each text is encoded into an unweighted and
undirected graph in which nodes represent selected terms and edges
represent co-occurrences between terms within a fixed sliding
window. A term is a group of words clustered based on their
part-of-speech tags such as noun and adjective. The choice of
sliding window size depends on the average lengths of processed
texts. Figure~\ref{fig:method} provides an example of this method.

\subsection{GCN and its Applications on Text Modeling}\label{sec:gcn}

Recently, many studies~\cite{kipf2016semi,velivckovic2017graph} have
attempted to apply convolutional operations on graph data. Graph
convolutional networks~(GCNs)~\cite{kipf2016semi} were proposed and
have achieved the state-of-the-art performance on inductive and
transductive node classification tasks. The spectral graph
convolution operation is defined such that a convolution-like
operation can be applied on graph data. Basically, each GCN layer
updates the feature representation of each node by aggregating the
features of its neighboring nodes. To be specific, the layer-wise
propagation rule of GCNs can be defined as
  $\mbox{gcn}(\hat{A}, X_{\ell}) =
  \sigma(\hat{D}^{-\frac{1}{2}}\hat{A}\hat{D}^{-\frac{1}{2}}X_{\ell}W_{\ell})$,
where $X_{\ell}$ and $X_{\ell+1}$ are input and output matrices of
layer $\ell$, respectively. The numbers of rows of two matrices are
the same, which indicates that the number of nodes in graph does not
change in GCN layers. We use $\hat{A} = A + I$
to include self-loop edges to the graph. The diagonal node degree
matrix $\hat{D}$ is generated to normalize $\hat{A}$ such that the
scale of feature vectors remains the same after aggregation.
$W^\ell$ is the trainable weight matrix, which plays the role of
linear transformation of each node's feature vector. Finally,
$\sigma(\cdot)$ denotes the nonlinear activation function such as
ReLU~\cite{nair2010rectified}. Later, graph attention
networks~(GAT)~\cite{velivckovic2017graph} are proposed to solve
node classification tasks by using the attention
mechanism~\cite{xu2015show}.

In addition to graph data, some studies attempted to apply
graph-based methods to grid-like data such as texts. Compared to
traditional recurrent neural networks such as
LSTM~\cite{hochreiter1997long}, GCNs have the advantage of
considering long-term dependencies by edges in graphs.
\citet{marcheggiani2017encoding} applied a variant of GCNs to the
task of sentence encoding and achieved better performance than LSTM.
GCNs have also been used in neural machine translation
tasks~\cite{bastings2017graph}. Although graph convolutional
operations have been extensively developed and explored, pooling
operations on graphs are not well studied currently.


\begin{figure*}[th] \centering
\includegraphics[width=\textwidth]{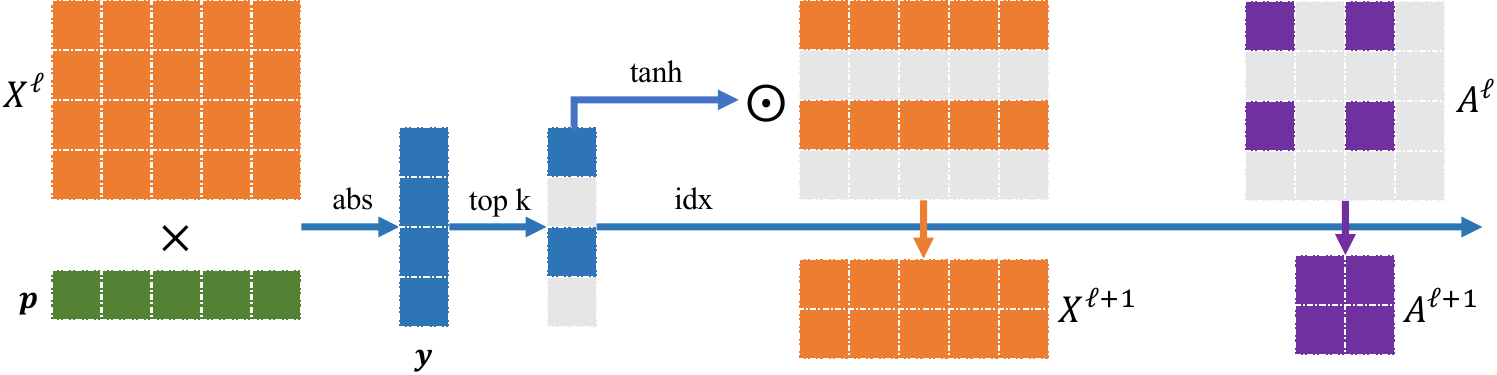}
\caption{An illustration of the proposed graph pooling layer which samples
$k=2$ nodes. $\times$ and $\odot$ represent matrix and element-wise
multiplication, respectively. This example graph has four nodes, each of which
contains 5 features. We have the adjacency matrix $A^\ell \in \mathbb{R}^{4
\times 4}$ and the input feature matrix $X^\ell \in \mathbb{R}^{4 \times 5} $
of layer $\ell$ representing this graph. $\mathbf p \in \mathbb{R}^{5} $ is a
trainable projection vector in this layer. By matrix multiplication and
$\mbox{abs}(\cdot)$, we have scores $\mathbf y$ which estimate the closeness
of each node to the projection vector. Using $k=2$, we select two nodes with
the highest scores and record their indices in $\mbox{idx}$, which represents
indices of selected nodes. With indices $\mbox{idx}$, we extract corresponding
nodes to form the new graph, which results in a new pooled feature map $\tilde
X^{\ell}$ and adjacency matrix $A^{\ell+1}$. To control information flow, we
create a gate vector by applying element-wise $\tanh(\cdot)$ to score vector $\mathbf y$. By element-wise
multiplication between gate vector and 
$\tilde X^{\ell}$, we obtain the $X^{\ell+1}$. Outputs of this
graph pooling layer are $A^{\ell+1}$ and $X^{\ell+1}$.}
\label{fig:GraphP}
\end{figure*}


\section{Graph Pooling and Hybrid Convolutional Operations}

In this section, we describe our proposed graph pooling~(gPool) and
hybrid convolutional~(hConv) operations. Based on these two new
layers, we develop FCN-like graph convolutional networks for text
modeling tasks.

\subsection{Graph Pooling Layer}\label{sec:gpool}

Pooling layers are very important for CNNs on grid-like data such as
images, as they can help quickly enlarge receptive fields and reduce
feature map size, thereby resulting in better generalization and
performance~\cite{yu2015multi}. In pooling layers, input feature
maps are partitioned into a set of non-overlapping rectangles on
which non-linear down-sampling functions, such as maximum, are
applied. Obviously, the partition scheme depends on the local
adjacency on grid-like data like images. For instance in a
max-pooling layer with a kernel size of $2 \times 2$, 4 spatially
adjacent units form a partition. However, such kind of spatial
adjacency information is not available for nodes on graphs. We
cannot directly apply regular pooling layers to graph data.

To enable pooling operations on graph data, we propose the graph
pooling layer~(gPool), which adaptively selects a subset of
important nodes to form a smaller new graph. Suppose a graph has $N$
nodes and each node contains $C$ features. We can represent the
graph with two matrices; those are the adjacency matrix $A^{\ell}
\in \mathbb{R}^{N \times N}$ and the feature matrix $X^{\ell} \in
\mathbb{R}^{N \times C}$. Each row in $X^{\ell}$ corresponds to a
node in the graph. We propose the layer-wise propagation rule of
gPool to be defined as
\begin{equation}
\begin{aligned}
  \mathbf y & = |X^{\ell} \mathbf p^{\ell}|,
  &\mbox{idx} & = \mbox{rank}(\mathbf y, k), \\
  A^{\ell+1} &= A^{\ell}(\mbox{idx}, \mbox{idx}),
  &\tilde X^{\ell} &= X^{\ell}(\mbox{idx}, :), \\
  \tilde{\mathbf y} &= \tanh(\mathbf y(\mbox{idx})),
  &X^{\ell+1} & = \tilde X^{\ell} \odot \left(\tilde{\mathbf y} \mathbf{1}_C^{T}\right),
\end{aligned}\label{eq:gpool}
\end{equation}
where $k$ is the number of nodes to be selected from the graph,
$\mathbf p^{\ell} \in \mathbb{R}^{C}$ is a trainable projection
vector of layer $\ell$, $\mbox{rank}(\mathbf y)$ is an operation
that returns the indices corresponding to the $k$-largest values in
$\mathbf y$, $A^{\ell}(\mbox{idx}, \mbox{idx})$ extracts the rows
and columns with indices idx, $X^{\ell}(\mbox{idx}, :)$ selects the
rows with indices idx using all column values, $\mathbf
y(\mbox{idx})$ returns the corresponding values in $\mathbf y$ with
indices idx, $\mathbf 1_{C}\in \mathbb{R}^{C}$ is a vector of size
$C$ with all components being 1, and $\odot$ denotes the
element-wise matrix multiplication operation.

Suppose $X^{\ell}$ is a matrix with row vectors $\mathbf x^{\ell}_1,
\mathbf x^{\ell}_2, \cdots, \mathbf x^{\ell}_N$. We first perform a
matrix multiplication between $X^{\ell}$ and $\mathbf{p}^{\ell}$
followed by an element-wise $\mbox{abs}(\cdot)$ operation, resulting in
$\mathbf y = [ y_1, y_2, \cdots, y_N ]^T$ with each $y_i$ measuring
the closeness between the node feature vector $\mathbf x^{\ell}_i$
and the projection vector $\mathbf p^{\ell}$.

The $\mbox{rank}(\cdot)$ operation ranks the $N$ values in $\mathbf y$ and
returns the indices of the $k$-largest values. Suppose the indices of the $k$
selected values are $i_1, i_2, \cdots, i_k $ with $i_m < i_n$ if $1 \le m < n
\le k$. These indices correspond to nodes in the graph. Note that the
selection process retains the order information of selected nodes in the
original feature matrix. With indices, the $k$-node selection is conducted on
the feature matrix $X^l$, the adjacency matrix $A^{\ell}$, and the score
vector $\mathbf y$. We concatenate the corresponding feature vectors $\mathbf
x^{\ell}_{i_1}, \mathbf x^{\ell}_{i_2}, \cdots, \mathbf x^{\ell}_{i_k}$ and
output a matrix $\tilde X^{\ell} \in \mathbb{R}^{k \times C}$. Similarly, we
use scores $\tanh(y_{i_1}), \tanh(y_{i_2}), \cdots, \tanh(y_{i_k})$ as
elements of the vector $\mathbf{\tilde y} \in \mathbb{R}^{k}$. For the
adjacency matrix $A^{\ell}$, we extract rows and columns based on the selected
indices and output the adjacency matrix $A^{{\ell}+1} \in \mathbb{R}^{k \times
k}$ for the new graph.

Finally, we perform a gate operation to control the information flow
in this layer. Using element-wise product of $\tilde X^{\ell}$ and
$\mathbf{\tilde y} \mathbf 1^T_C$, features of selected nodes are
filtered through their corresponding scores and form the output
$X^{{\ell}+1} \in \mathbb{R}^{k \times C}$. The $i$th row vector in
$X^{{\ell}+1}$ is the product of the corresponding row vector in
$\tilde X^{\ell}$ and the $i^{th}$ scalar number in $\mathbf{\tilde
y}$. In addition to information control, the gate operation also
makes the projection vector $\mathbf p^{\ell}$ trainable with
back-propagation~\cite{lecun2012efficient}. Without the gate operation,
the projection vector $\mathbf p^{\ell}$ only contributes discrete
indices to outputs and thus is not trainable.
Figure~\ref{fig:GraphP} provides an illustration of the gPool layer.

Compared to regular pooling layers used on grid-like data, our gPool
layer involves extra parameters in the trainable projection vector
$\mathbf p^{\ell}$. In Section~\ref{sec:params}, we show that the
number of extra parameters is negligible and would not increase the
risk of over-fitting.

\subsection{Hybrid Convolutional Layer}\label{sec:hconv}

It follows from the analysis in Section~\ref{sec:gcn} that GCN
layers only perform convolutional operations on each node. There is
no trainable spatial filters as in regular convolution layers. GCNs
do not have the power of automatic feature extraction as achieved by
CNNs. This limits the capability of GCNs, especially in the field of
graph modeling. In traditional graph data, there is no ordering
information among nodes. In addition, the different numbers of
neighbors for each node in the graph prohibit convolutional
operations with a kernel size larger than 1. Although we attempt to
modeling texts as graph data, they are essentially grid-like data
with order information among nodes, thereby enabling the application
of regular convolutional operations.

\begin{figure}[t] \includegraphics[width=0.8\columnwidth]{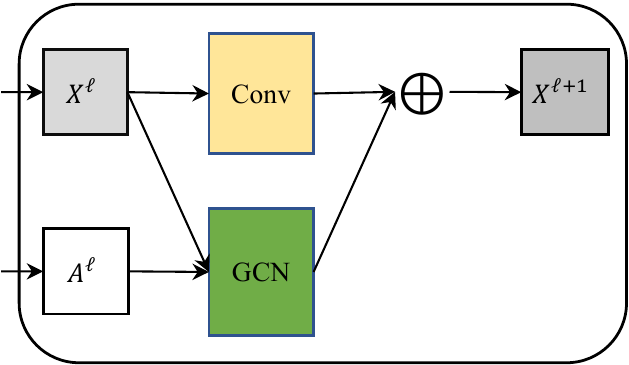}
\caption{An illustration of the hybrid convolutional layer. $\oplus$
denotes matrix concatenation along the row dimension. In this layer,
$A^\ell$ and $X^\ell$ are the adjacency matrix and the node feature
matrix, respectively. A regular 1-D convolutional operation is used
to extract high-level features from sentence texts. The GCN
operation is applied at the graph level for feature extraction. The
two intermediate outputs are concatenated together to form the final
output $X^{\ell+1}$.} \label{fig:hconv} \end{figure}

To take advantage of convolutional operations with trainable
filters, we propose the hybrid convolutional layer~(hConv), which
combines GCN operations and regular 1-D convolutional operations to
achieve the capability of automatic feature extraction. Formally, we
propose the hConv layer to be defined as
\begin{equation}
\begin{aligned}
  X^{\ell+1}_{1} & = \mbox{conv}(X^\ell), 
  &X^{\ell+1}_{2} & = \mbox{gcn}(A^\ell, X^\ell), \\
  X^{\ell+1} & = \big[ X^{\ell+1}_{1}, X^{\ell+1}_{2} \big],
\end{aligned}\label{eq:hconv}
\end{equation}
where $\mbox{conv}(\cdot)$ denotes a regular 1-D convolutional
operation, and the $\mbox{gcn}(\cdot, \cdot)$ operation is defined
in Eq.~\ref{eq:gcn}. For the feature matrix $X^\ell$, we treat the
column dimension as the channel dimension, such that the 1-D
convolutional operation can be applied along the row dimension.
Using the 1-D convolutional operation and the GCN operation, we
obtain two intermediate outputs; those are $X^{\ell+1}_{1}$ and
$X^{\ell+1}_{2}$. These two matrices are concatenated together as
the layer output $X^{\ell+1}$. Figure~\ref{fig:hconv} illustrates an
example of the hConv layer.

We argue that the integration of GCN operations and 1-D
convolutional operations in the hConv layer is especially applicable
to graph data obtained from texts. By representing texts as an
adjacency matrix $A^\ell$ and a node feature matrix $X^\ell$ of
layer $\ell$, each node in the graph is essentially a word in the
text. We retain the order information of nodes from their original
relative positions in texts. This indicates that the feature matrix
$X^{\ell}$ is organized as traditional grid-like data with order
information retained. From this point, we can apply an 1-D
convolutional operation with kernel sizes larger than 1 on the
feature matrix $X^\ell$ for high-level feature extraction.

The combination of the GCN operation and the convolutional operation
in the hConv layer can take the advantages of both of them and
overcome their respective limitations. In convolutional layers, the
receptive fields of units on feature maps increase very slow since
small kernel sizes are usually used to avoid massive number of
parameters. In contrast, GCN operations can help to increase the
receptive fields quickly by means of edges between terms in
sentences corresponding to nodes in graphs. At the same time, GCN
operations are not able to automatically extract high-level features
as they do not have trainable spatial filters as used in
convolutional operations. From this point, the hConv layer is
especially useful when working on text-based graph data such as
sentences and documents.

\begin{figure*}[t] \centering
\includegraphics[width=0.95\textwidth]{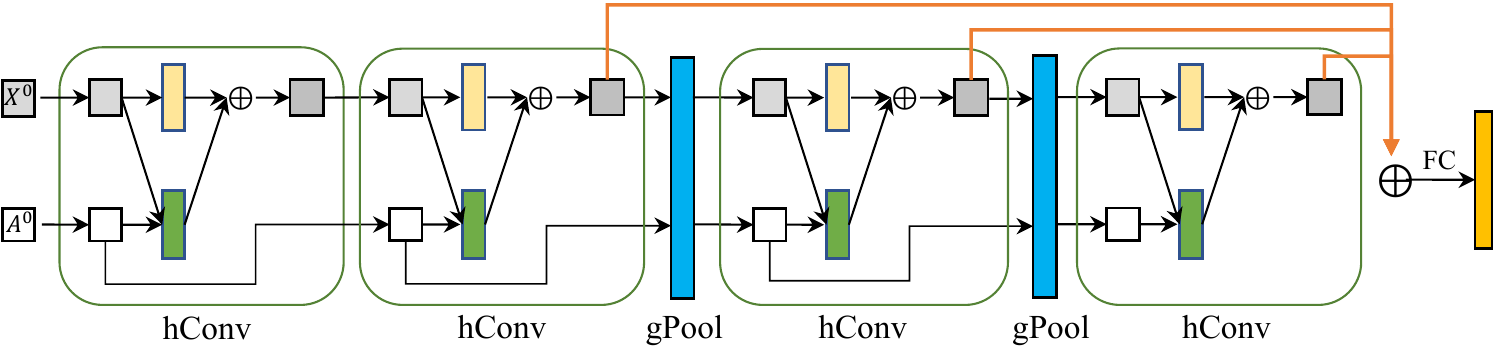}
\caption{The architecture of the hConv-gPool-Net. $\oplus$ denotes
the concatenation operation of feature maps. The inputs of the
network are an adjacency matrix $A^0$ and a feature matrix $X^0$. We
stack four hConv layers for feature extraction. In the second and
the third hConv layers, we employ the gPool layers to reduce the
number of nodes in graphs by half. Starting from the second hConv
layer, a global max-pooling layer is applied to the output feature
maps of each hConv layer. The outputs of these pooling layers are
concatenated together. Finally, we employ a fully-connected layer
for predictions. To obtain the other three networks discussed in
Section~\ref{sec:arch}, we can simply replace the hConv layers with
GCN layers or remove gPool layers based on this network
architecture.} \label{fig:architecture} \end{figure*}

\subsection{Network Architectures}\label{sec:arch}

Based on our proposed gPool and hConv layers, we design four network
architectures, including our baseline with a
FCN-like~\cite{long2015fully} architecture. FCN has been shown to be
very effective for image semantic segmentation. It allows final
linear classifiers to make use of features from different layers.
Here, we design four architectures based on our proposed gPool and
hConv layers.

\begin{itemize}[leftmargin=*]
  \item \textbf{GCN-Net:} We establish a baseline method by using GCN layers to build a network
without any hConv or gPool layers. In this network, we stack 4
standard GCN layers as feature extractors. Starting from the second
layer, a global max-pooling layer~\cite{lin2013network} is applied
to each layer's output. The outputs of these pooling layers are
concatenated together and fed into a fully-connected layer for final
predictions. This network serves as a baseline model in this work
for experimental studies.

  \item \textbf{GCN-gPool-Net:} In this network, we add our proposed gPool
layers to GCN-Net. Starting from the second layer, we add a gPool
layer after each GCN layer except for the last one. In each gPool
layer, we select the hyper-parameter $k$ to reduce the number of
nodes in the graph by a factor of two. All other parts of the
network remain the same as those of GCN-Net.

  \item \textbf{hConv-Net:} For this network, we replace all GCN layers in
GCN-Net by our proposed hConv layers. To ensure the fairness of
comparison among these networks, the hConv layers output the same
number of feature maps as the corresponding GCN layers. Suppose the
original $i$th GCN layer outputs $n_{out}$ feature maps. In the
corresponding hConv layer, both the GCN operation and the
convolutional operation output $n_{out}/2$ feature maps. By
concatenating those intermediate outputs, the $i$th hConv layer also
outputs $n_{out}$ feature maps. The remaining parts of the network
remain the same as those in GCN-Net.

  \item \textbf{hConv-gPool-Net:} Based on the hConv-Net network, we add gPool
layers after each hConv layer except for the first and the last
layers. We employ the same principle for the selection of
hyper-parameter $k$ as that in GCN-gPool-Net. The remaining parts of
the network remain the same. Note that gPool layers maintain the
order information of nodes in the new graph, thus enabling the
application of 1-D convolutional operations in hConv layers
afterwards. Figure~\ref{fig:architecture} provides an illustration
of the hConv-gPool-Net network.
\end{itemize}

\begin{table}[t]
\begin{tabularx}{\columnwidth}{ lx{1.8cm} YYYY }\hline
\bf Datasets & \#\bf Training & \#\bf Testing & \#\bf Classes &
\#\bf Words \\ \hline\hline
AG's~News     & 120,000  & 7,600  & 4  & 45 \\
DBPedia       & 560,000  & 70,000 & 14 & 55 \\
Yelp~Polarity & 560,000  & 38,000 & 2  & 153 \\
Yelp~Full    & 650,000   & 50,000 & 5  & 155 \\\hline
\end{tabularx}
\caption{Summary of datasets used in our experiments. The \#words
denotes the average number of words of the data samples for each
dataset. These numbers help the selection of the sliding window size
used in converting texts to graphs.} \label{table:dataset}
\end{table}

\begin{table*}[t]
\begin{tabularx}{\textwidth}{X YYYY }
\hline \bf Models          & \bf AG's News        & \bf DBPedia   &
\bf Yelp Polarity   & \bf Yelp Full\\ \hline \hline
Word-level CNN~\cite{zhang2015character}  & 8.55\%     & 1.37\%      & 4.60\%     & 39.58\% \\
Char-level CNN~\cite{zhang2015character}    & 9.51\%      & 1.55\% &
4.88\%     & 37.95\% \\\hline
GCN-Net              & 8.64\%     & 1.69\%      & 7.74\%     & 42.60\% \\
GCN-gPool-Net        & 8.09\%     & 1.44\%      & 5.82\%        & 41.83\% \\
hConv-Net            & 7.49\%     & 1.02\%      & 4.45\%        & 37.81\% \\
hConv-gPool-Net      & \bf 7.09\% & \bf 0.92\%  & \bf 4.37\% & \bf
36.27\%\\\hline
\end{tabularx}
\caption{Results of text classification experiments in terms of
classification error rate on the AG's News, DBPedia, Yelp Review
Polarity, and Yelp Review Full datasets. The first two methods are
the state-of-the-art models without using any unsupervised data. The
last four networks are proposed in this work.} \label{table:results}
\end{table*}


\section{Experimental Study}

In this section, we evaluate our gPool layer and hConv layer based
on the four networks proposed in Section~\ref{sec:arch}. We compare
the performances of our networks with that of previous
state-of-the-art models. The experimental results show that our
methods yield improved performance in terms of classification
accuracy. We also perform some ablation studies to examine the
contributions of the gPool layer and the hConv layer to the
performance. The number of extra parameters in gPool layers is shown
to be negligible and will not increase the risk of over-fitting.

\subsection{Datasets}

In this work, we evaluate our methods on four datasets, including
the AG's News, Dbpedia, Yelp Polarity, and Yelp
Full~\cite{zhang2015character} datasets.

\textbf{AG's News} is a news dataset containing four topics: World,
Sports, Business and Sci/Tech. The task is to classify each news
into one of the topics.

\textbf{Dbpedia} ontology dataset contains 14 ontology classes. It
is constructed by choosing 14 non-overlapping classes from the
DBPedia 2014 dataset~\cite{lehmann2015dbpedia}. Each sample contains
a title and an abstract corresponding to a Wikipedia article.

\textbf{Yelp Polarity} dataset is obtained from the Yelp Dataset
Challenge in 2015~\cite{zhang2015character}. Each sample is a piece of review
text with a binary label (negative or positive).

\textbf{Yelp Full} dataset is obtained from the Yelp Dataset Challenge in
2015, which is for sentiment classification~\cite{zhang2015character}. It
contains five classes corresponding to the movie review star ranging from 1 to
5.

The summary of these datasets are provided in
Table~\ref{table:dataset}. For all datasets, we tokenize the textual
document and convert words to lower case. We remove stop-words and
all punctuation in texts. Based on cleaned texts, we build the
graph-of-word representations for texts.

\subsection{Text to Graph Conversion}

We use the graph-of-words method to convert texts into graph
representations that include an adjacency matrix and a feature
matrix. We select nouns, adjective, and verb as terms, meaning a
word appears in the graph if it belongs to one of the above
categories. We use a sliding window to decide if two terms have an
edge between them. If the distance between two terms is less than
the window size, an undirected edge between these two terms is
added. In the generated graph, nodes are the terms appear in texts,
and edges are added using the sliding window. We use a window size
of 4 for the AG's News and DBpedia datasets and 10 for the other two
datasets, depending on their average words in training samples. The
maximum numbers of nodes in graphs for the AG's News, DBPedia, Yelp
Polarity, Yelp Full datasets are 100, 100, 300, and 256,
respectively.

To produce the feature matrix, we use word embedding and position
embedding features. For word embedding features, the pre-trained
fastText word embedding vectors~\cite{joulin2017bag} are used, and
it contains more than 2 million pre-trained words vectors. Compared
to other pre-trained word embedding vectors such as
GloVe~\cite{pennington2014glove}, using the fastText helps us to
avoid massive unknown words. On the AG's News dataset, the number of
unknown words with the fastText is only several hundred, which is
significantly smaller than the number using GloVe. In addition to
word embedding features, we also employ position embedding method
proposed in~\citet{zeng2014relation}. We encode the positions of
words in texts into one-hot vectors and concatenate them with word
embedding vectors. We obtain the feature matrix by stacking word
vectors of nodes in the row dimension.

\begin{table}[t]
\begin{tabularx}{\columnwidth}{lx{3.2cm} YY}
\hline \bf Models     & \bf Depth & \bf Error Rate \\ \hline \hline
Word-level CNN     & 9 & 8.55\% \\
Character-level CNN     & 9 & 9.51\% \\\hline
GCN-Net            & 5 & 8.64\%    \\
GCN-gPool-Net      & 5 & 8.09\% \\
hConv-Net          & 5 & 7.49\%    \\
hConv-gPool-Net    & 5 & \textbf{7.09\%} \\
\hline
\end{tabularx}
\caption{\label{table:depth} Comparison in terms of the network
depth and the text classification error rate on the AG's News
dataset. The depth listed here is calculated by counting the number
of convolutional and fully-connected layers in networks.}
\end{table}

\subsection{Experimental Setup}

For our proposed networks, we employ the same settings with minor
adjustments to accommodate the different datasets. As discussed in
Section~\ref{sec:arch}, we stack four GCN or hConv layers for
GCN-based networks or hConv-based networks. For the networks using
gPool layers, we add gPool layers after the second and the third GCN
or hConv layers. Four GCN or hConv layers output 1024, 1024, 512,
and 256 feature maps, respectively. We use this decreasing number of
feature maps, since GCNs help to enlarge the receptive fields very
quickly. We do not need more high-level features in deeper layers.
The kernel sizes used by convolutional operations in hConv layers
are all $3 \times 1$. For all layers, we use the
ReLU~\cite{nair2010rectified} for nonlinearity.
For all experiments, the following settings are shared. For
training, the Adam optimizer~\cite{kingma2014adam} is used for 60
epochs. The learning rate starts at 0.001 and decays by 0.1 at the
30$^th$ and the 50$^th$ epoch. We employ the dropout with a keep
rate of 0.55~\cite{srivastava2014dropout} and batch size of 256.
These hyper-parameters are tuned on the AG's News dataset, and then
ported to other datasets.

\subsection{Performance Study}

We compare our proposed methods with other state-of-the-art models,
and the experimental results are summarized in
Table~\ref{table:results}. We can see from the results that our
hConv-gPool-Net outperforms both word-level CNN and character-level
CNN by at least a margin of 1.46\%, 0.45\%, 0.23\%, and 3.31\% on
the AG's News, DBPedia, Yelp Polarity, and Yelp Full datasets,
respectively. The performance of GCN-Net with only GCN layers cannot
compete with that of word-level CNN and char-level CNN primarily due
to the lack of automatic high-level feature extraction. By replacing
GCN layers using our proposed hConv layers, hConv-Net achieves
better performance than the two CNN models across four datasets.
This demonstrates the promising performance of our hConv layer by
employing regular convolutional operations for automatic feature
extraction.
By comparing the GCN-Net with GCN-gPool-Net, and hConv-Net with
hConv-gPool-Net, we observe that our proposed gPool layers promote
both models' performance by at least a margin of 0.4\%, 0.1\%,
0.08\%, and 1.54\% on the AG's News, DBPedia, Yelp Polarity, and
Yelp Full datasets. The margins tend to be larger on harder tasks.
This observation demonstrates that our gPool layer helps to enlarge
receptive fields and reduce spatial dimensions of graphs, resulting
in better generalization and performance.

\begin{table}[t]
\begin{tabularx}{\columnwidth}{l c c c}
\hline \bf Models     & \bf Error rate    & \# \bf Params & \bf Ratio of increase\\ \hline\hline
GCN-Net        & 8.64\%            & 1,554,820     & 0.00\%\\
GCN-gPool-Net         & \textbf{8.09\%}   & 1,555,719     & 0.06\%\\
\hline
\end{tabularx}
\caption{\label{table:params} Comparison between the GCN-Net and
GCN-gPool-Net in term of parameter numbers and text classification
error rates on the AG's News dataset.}
\end{table}

\subsection{Network Depth Study}

In addition to performance study, we also conduct experiments to
evaluate the relationship between performance and network depth in
terms of the number of convolutional and fully-connected layers in
models. The results are summarized in Table~\ref{table:depth}. We
can observe from the results that our models only require 5 layers,
including 4 convolutional layers and 1 fully-connected layer. Both
word-level CNN and character-level CNN models need 9 layers in their
networks, which are much deeper than ours. Our hConv-gPool-Net
achieves the new state-of-the-art performance with fewer layers,
demonstrating the effectiveness of gPool and hConv layers. Since
GCN and gPool layers enlarge receptive fields quickly,
advanced features are learned in shallow layers, leading to shallow
networks but better performance.

\subsection{Parameter Study of gPool Layer}\label{sec:params}

Since gPool layers involve extra trainable parameters in projection
vectors, we study the number of parameters in gPool layers in the
GCN-gPool-Net that contains two gPool layers. The results are
summarized in Table~\ref{table:params}. We can see from the results
that gPool layers only needs 0.06\% additional parameters compared
to GCN-Net. We believe that this negligible increase of parameters
will not increase the risk of over-fitting. With negligible
additional parameters, gPool layers can yield a performance
improvement of 0.54\%.

\section{Conclusion}

In this work, we propose the gPool and hConv layers in FCN-like
graph convolutional networks for text modeling. The gPool layer
achieves the effect of regular pooling operations on graph data to
extract important nodes in graphs. By learning a projection vector,
all nodes are measured through cosine similarity with the projection
vector. The nodes with the $k$-largest scores are extracted to form
a new graph. The scores are then applied to the feature matrix for
information control, leading to the additional benefit of making the
projection vector trainable. Since graphs are extracted from texts,
we maintain the node orders as in the original texts. We propose the
hConv layer that combines GCN and regular convolutional operations
to enable automatic high-level feature extraction. Based on our
gPool and hConv layers, we propose four networks for the task of
text categorization. Our results show that the model based on gPool
and hConv layers achieves new state-of-the-art performance compared
to CNN-based models. gPool layers involve negligible number of
parameters but bring significant performance boosts, demonstrating
its contributions to model performance.

\section*{Acknowledgments}

This work was supported in part by National Science Foundation grant
IIS-1908166.

\newpage

\bibliographystyle{ACM-Reference-Format}
\balance 
\bibliography{nlp}


\begin{thebibliography}{38}


\ifx \showCODEN    \undefined \def \showCODEN     #1{\unskip}     \fi
\ifx \showDOI      \undefined \def \showDOI       #1{#1}\fi
\ifx \showISBNx    \undefined \def \showISBNx     #1{\unskip}     \fi
\ifx \showISBNxiii \undefined \def \showISBNxiii  #1{\unskip}     \fi
\ifx \showISSN     \undefined \def \showISSN      #1{\unskip}     \fi
\ifx \showLCCN     \undefined \def \showLCCN      #1{\unskip}     \fi
\ifx \shownote     \undefined \def \shownote      #1{#1}          \fi
\ifx \showarticletitle \undefined \def \showarticletitle #1{#1}   \fi
\ifx \showURL      \undefined \def \showURL       {\relax}        \fi
\providecommand\bibfield[2]{#2}
\providecommand\bibinfo[2]{#2}
\providecommand\natexlab[1]{#1}
\providecommand\showeprint[2][]{arXiv:#2}

\bibitem[\protect\citeauthoryear{Bahdanau, Cho, and Bengio}{Bahdanau
  et~al\mbox{.}}{2015}]%
        {bahdanau2014neural}
\bibfield{author}{\bibinfo{person}{Dzmitry Bahdanau},
  \bibinfo{person}{Kyunghyun Cho}, {and} \bibinfo{person}{Yoshua Bengio}.}
  \bibinfo{year}{2015}\natexlab{}.
\newblock \showarticletitle{Neural machine translation by jointly learning to
  align and translate}.
\newblock \bibinfo{journal}{\emph{International Conference on Learning
  Representations}} (\bibinfo{year}{2015}).
\newblock


\bibitem[\protect\citeauthoryear{Bastings, Titov, Aziz, Marcheggiani, and
  Sima'an}{Bastings et~al\mbox{.}}{2017}]%
        {bastings2017graph}
\bibfield{author}{\bibinfo{person}{Joost Bastings}, \bibinfo{person}{Ivan
  Titov}, \bibinfo{person}{Wilker Aziz}, \bibinfo{person}{Diego Marcheggiani},
  {and} \bibinfo{person}{Khalil Sima'an}.} \bibinfo{year}{2017}\natexlab{}.
\newblock \showarticletitle{Graph convolutional encoders for syntax-aware
  neural machine translation}.
\newblock \bibinfo{journal}{\emph{arXiv preprint arXiv:1704.04675}}
  (\bibinfo{year}{2017}).
\newblock


\bibitem[\protect\citeauthoryear{Bronselaer and Pasi}{Bronselaer and
  Pasi}{2013}]%
        {bronselaer2013approach}
\bibfield{author}{\bibinfo{person}{Antoon Bronselaer} {and}
  \bibinfo{person}{Gabriella Pasi}.} \bibinfo{year}{2013}\natexlab{}.
\newblock \showarticletitle{An approach to graph-based analysis of textual
  documents}. In \bibinfo{booktitle}{\emph{8th European Society for Fuzzy Logic
  and Technology (EUSFLAT-2013)}}. Atlantis Press, \bibinfo{pages}{634--641}.
\newblock


\bibitem[\protect\citeauthoryear{He, Zhang, Ren, and Sun}{He
  et~al\mbox{.}}{2016}]%
        {he2015deep}
\bibfield{author}{\bibinfo{person}{Kaiming He}, \bibinfo{person}{Xiangyu
  Zhang}, \bibinfo{person}{Shaoqing Ren}, {and} \bibinfo{person}{Jian Sun}.}
  \bibinfo{year}{2016}\natexlab{}.
\newblock \showarticletitle{Deep Residual Learning for Image Recognition}. In
  \bibinfo{booktitle}{\emph{Proceedings of the IEEE Conference on Computer
  Vision and Pattern Recognition}}.
\newblock


\bibitem[\protect\citeauthoryear{Hochreiter and Schmidhuber}{Hochreiter and
  Schmidhuber}{1997}]%
        {hochreiter1997long}
\bibfield{author}{\bibinfo{person}{Sepp Hochreiter} {and}
  \bibinfo{person}{J{\"u}rgen Schmidhuber}.} \bibinfo{year}{1997}\natexlab{}.
\newblock \showarticletitle{Long short-term memory}.
\newblock \bibinfo{journal}{\emph{Neural computation}} \bibinfo{volume}{9},
  \bibinfo{number}{8} (\bibinfo{year}{1997}), \bibinfo{pages}{1735--1780}.
\newblock


\bibitem[\protect\citeauthoryear{Huang, Liu, Weinberger, and van~der
  Maaten}{Huang et~al\mbox{.}}{2017}]%
        {huang2017densely}
\bibfield{author}{\bibinfo{person}{Gao Huang}, \bibinfo{person}{Zhuang Liu},
  \bibinfo{person}{Kilian~Q Weinberger}, {and} \bibinfo{person}{Laurens van~der
  Maaten}.} \bibinfo{year}{2017}\natexlab{}.
\newblock \showarticletitle{Densely connected convolutional networks}. In
  \bibinfo{booktitle}{\emph{Proceedings of the IEEE conference on computer
  vision and pattern recognition}}, Vol.~\bibinfo{volume}{1}.
  \bibinfo{pages}{3}.
\newblock


\bibitem[\protect\citeauthoryear{Joulin, Grave, Bojanowski, and Mikolov}{Joulin
  et~al\mbox{.}}{2017}]%
        {joulin2017bag}
\bibfield{author}{\bibinfo{person}{Armand Joulin}, \bibinfo{person}{Edouard
  Grave}, \bibinfo{person}{Piotr Bojanowski}, {and} \bibinfo{person}{Tomas
  Mikolov}.} \bibinfo{year}{2017}\natexlab{}.
\newblock \showarticletitle{Bag of Tricks for Efficient Text Classification}.
  In \bibinfo{booktitle}{\emph{Proceedings of the 15th Conference of the
  European Chapter of the Association for Computational Linguistics: Volume 2,
  Short Papers}}, Vol.~\bibinfo{volume}{2}. \bibinfo{pages}{427--431}.
\newblock


\bibitem[\protect\citeauthoryear{Kingma and Ba}{Kingma and Ba}{2015}]%
        {kingma2014adam}
\bibfield{author}{\bibinfo{person}{Diederik Kingma} {and}
  \bibinfo{person}{Jimmy Ba}.} \bibinfo{year}{2015}\natexlab{}.
\newblock \showarticletitle{Adam: A method for stochastic optimization}.
\newblock \bibinfo{journal}{\emph{The International Conference on Learning
  Representations}} (\bibinfo{year}{2015}).
\newblock


\bibitem[\protect\citeauthoryear{Kipf and Welling}{Kipf and Welling}{2017}]%
        {kipf2016semi}
\bibfield{author}{\bibinfo{person}{Thomas~N Kipf} {and} \bibinfo{person}{Max
  Welling}.} \bibinfo{year}{2017}\natexlab{}.
\newblock \showarticletitle{Semi-supervised classification with graph
  convolutional networks}.
\newblock \bibinfo{journal}{\emph{International Conference on Learning
  Representations}} (\bibinfo{year}{2017}).
\newblock


\bibitem[\protect\citeauthoryear{Krizhevsky, Sutskever, and Hinton}{Krizhevsky
  et~al\mbox{.}}{2012}]%
        {krizhevsky2012imagenet}
\bibfield{author}{\bibinfo{person}{Alex Krizhevsky}, \bibinfo{person}{Ilya
  Sutskever}, {and} \bibinfo{person}{Geoffrey~E Hinton}.}
  \bibinfo{year}{2012}\natexlab{}.
\newblock \showarticletitle{Imagenet classification with deep convolutional
  neural networks}. In \bibinfo{booktitle}{\emph{Advances in neural information
  processing systems}}. \bibinfo{pages}{1097--1105}.
\newblock


\bibitem[\protect\citeauthoryear{LeCun, Bottou, Bengio, and Haffner}{LeCun
  et~al\mbox{.}}{1998}]%
        {lecun1998gradient}
\bibfield{author}{\bibinfo{person}{Yann LeCun}, \bibinfo{person}{L{\'e}on
  Bottou}, \bibinfo{person}{Yoshua Bengio}, {and} \bibinfo{person}{Patrick
  Haffner}.} \bibinfo{year}{1998}\natexlab{}.
\newblock \showarticletitle{Gradient-based learning applied to document
  recognition}.
\newblock \bibinfo{journal}{\emph{Proc. IEEE}} \bibinfo{volume}{86},
  \bibinfo{number}{11} (\bibinfo{year}{1998}), \bibinfo{pages}{2278--2324}.
\newblock


\bibitem[\protect\citeauthoryear{LeCun, Bottou, Orr, and M{\"u}ller}{LeCun
  et~al\mbox{.}}{2012}]%
        {lecun2012efficient}
\bibfield{author}{\bibinfo{person}{Yann LeCun}, \bibinfo{person}{L{\'e}on
  Bottou}, \bibinfo{person}{Genevieve~B Orr}, {and}
  \bibinfo{person}{Klaus-Robert M{\"u}ller}.} \bibinfo{year}{2012}\natexlab{}.
\newblock \showarticletitle{Efficient backprop}.
\newblock In \bibinfo{booktitle}{\emph{Neural networks: Tricks of the trade}}.
  \bibinfo{publisher}{Springer}, \bibinfo{pages}{9--48}.
\newblock


\bibitem[\protect\citeauthoryear{Lehmann, Isele, Jakob, Jentzsch, Kontokostas,
  Mendes, Hellmann, Morsey, Van~Kleef, Auer, et~al\mbox{.}}{Lehmann
  et~al\mbox{.}}{2015}]%
        {lehmann2015dbpedia}
\bibfield{author}{\bibinfo{person}{Jens Lehmann}, \bibinfo{person}{Robert
  Isele}, \bibinfo{person}{Max Jakob}, \bibinfo{person}{Anja Jentzsch},
  \bibinfo{person}{Dimitris Kontokostas}, \bibinfo{person}{Pablo~N Mendes},
  \bibinfo{person}{Sebastian Hellmann}, \bibinfo{person}{Mohamed Morsey},
  \bibinfo{person}{Patrick Van~Kleef}, \bibinfo{person}{S{\"o}ren Auer},
  {et~al\mbox{.}}} \bibinfo{year}{2015}\natexlab{}.
\newblock \showarticletitle{DBpedia--a large-scale, multilingual knowledge base
  extracted from Wikipedia}.
\newblock \bibinfo{journal}{\emph{Semantic Web}} \bibinfo{volume}{6},
  \bibinfo{number}{2} (\bibinfo{year}{2015}), \bibinfo{pages}{167--195}.
\newblock


\bibitem[\protect\citeauthoryear{Lin, Chen, and Yan}{Lin et~al\mbox{.}}{2013}]%
        {lin2013network}
\bibfield{author}{\bibinfo{person}{Min Lin}, \bibinfo{person}{Qiang Chen},
  {and} \bibinfo{person}{Shuicheng Yan}.} \bibinfo{year}{2013}\natexlab{}.
\newblock \showarticletitle{Network in network}.
\newblock \bibinfo{journal}{\emph{arXiv preprint arXiv:1312.4400}}
  (\bibinfo{year}{2013}).
\newblock


\bibitem[\protect\citeauthoryear{Liu, Zhang, Niu, Lin, Lai, and Xu}{Liu
  et~al\mbox{.}}{2018}]%
        {liu2018matching}
\bibfield{author}{\bibinfo{person}{Bang Liu}, \bibinfo{person}{Ting Zhang},
  \bibinfo{person}{Di Niu}, \bibinfo{person}{Jinghong Lin},
  \bibinfo{person}{Kunfeng Lai}, {and} \bibinfo{person}{Yu Xu}.}
  \bibinfo{year}{2018}\natexlab{}.
\newblock \showarticletitle{Matching Long Text Documents via Graph
  Convolutional Networks}.
\newblock \bibinfo{journal}{\emph{arXiv preprint arXiv:1802.07459}}
  (\bibinfo{year}{2018}).
\newblock


\bibitem[\protect\citeauthoryear{Long, Shelhamer, and Darrell}{Long
  et~al\mbox{.}}{2015}]%
        {long2015fully}
\bibfield{author}{\bibinfo{person}{Jonathan Long}, \bibinfo{person}{Evan
  Shelhamer}, {and} \bibinfo{person}{Trevor Darrell}.}
  \bibinfo{year}{2015}\natexlab{}.
\newblock \showarticletitle{Fully convolutional networks for semantic
  segmentation}. In \bibinfo{booktitle}{\emph{Proceedings of the IEEE
  conference on computer vision and pattern recognition}}.
  \bibinfo{pages}{3431--3440}.
\newblock


\bibitem[\protect\citeauthoryear{Malliaros and Skianis}{Malliaros and
  Skianis}{2015}]%
        {malliaros2015graph}
\bibfield{author}{\bibinfo{person}{Fragkiskos~D Malliaros} {and}
  \bibinfo{person}{Konstantinos Skianis}.} \bibinfo{year}{2015}\natexlab{}.
\newblock \showarticletitle{Graph-based term weighting for text
  categorization}. In \bibinfo{booktitle}{\emph{Advances in Social Networks
  Analysis and Mining (ASONAM), 2015 IEEE/ACM International Conference on}}.
  IEEE, \bibinfo{pages}{1473--1479}.
\newblock


\bibitem[\protect\citeauthoryear{Marcheggiani and Titov}{Marcheggiani and
  Titov}{2017}]%
        {marcheggiani2017encoding}
\bibfield{author}{\bibinfo{person}{Diego Marcheggiani} {and}
  \bibinfo{person}{Ivan Titov}.} \bibinfo{year}{2017}\natexlab{}.
\newblock \showarticletitle{Encoding sentences with graph convolutional
  networks for semantic role labeling}.
\newblock \bibinfo{journal}{\emph{arXiv preprint arXiv:1703.04826}}
  (\bibinfo{year}{2017}).
\newblock


\bibitem[\protect\citeauthoryear{Mihalcea and Tarau}{Mihalcea and
  Tarau}{2004}]%
        {mihalcea2004textrank}
\bibfield{author}{\bibinfo{person}{Rada Mihalcea} {and} \bibinfo{person}{Paul
  Tarau}.} \bibinfo{year}{2004}\natexlab{}.
\newblock \showarticletitle{Textrank: Bringing order into text}. In
  \bibinfo{booktitle}{\emph{Proceedings of the 2004 conference on empirical
  methods in natural language processing}}.
\newblock


\bibitem[\protect\citeauthoryear{Nair and Hinton}{Nair and Hinton}{2010}]%
        {nair2010rectified}
\bibfield{author}{\bibinfo{person}{Vinod Nair} {and}
  \bibinfo{person}{Geoffrey~E Hinton}.} \bibinfo{year}{2010}\natexlab{}.
\newblock \showarticletitle{Rectified linear units improve restricted boltzmann
  machines}. In \bibinfo{booktitle}{\emph{Proceedings of the 27th international
  conference on machine learning (ICML-10)}}. \bibinfo{pages}{807--814}.
\newblock


\bibitem[\protect\citeauthoryear{Niepert, Ahmed, and Kutzkov}{Niepert
  et~al\mbox{.}}{2016}]%
        {niepert2016learning}
\bibfield{author}{\bibinfo{person}{Mathias Niepert}, \bibinfo{person}{Mohamed
  Ahmed}, {and} \bibinfo{person}{Konstantin Kutzkov}.}
  \bibinfo{year}{2016}\natexlab{}.
\newblock \showarticletitle{Learning convolutional neural networks for graphs}.
  In \bibinfo{booktitle}{\emph{International Conference on Machine Learning}}.
  \bibinfo{pages}{2014--2023}.
\newblock


\bibitem[\protect\citeauthoryear{Nikolentzos, Meladianos, Rousseau, Stavrakas,
  and Vazirgiannis}{Nikolentzos et~al\mbox{.}}{2017}]%
        {nikolentzos2017shortest}
\bibfield{author}{\bibinfo{person}{Giannis Nikolentzos},
  \bibinfo{person}{Polykarpos Meladianos}, \bibinfo{person}{Fran{\c{c}}ois
  Rousseau}, \bibinfo{person}{Yannis Stavrakas}, {and}
  \bibinfo{person}{Michalis Vazirgiannis}.} \bibinfo{year}{2017}\natexlab{}.
\newblock \showarticletitle{Shortest-Path Graph Kernels for Document
  Similarity}. In \bibinfo{booktitle}{\emph{Proceedings of the 2017 Conference
  on Empirical Methods in Natural Language Processing}}.
  \bibinfo{pages}{1890--1900}.
\newblock


\bibitem[\protect\citeauthoryear{Pennington, Socher, and Manning}{Pennington
  et~al\mbox{.}}{2014}]%
        {pennington2014glove}
\bibfield{author}{\bibinfo{person}{Jeffrey Pennington},
  \bibinfo{person}{Richard Socher}, {and} \bibinfo{person}{Christopher
  Manning}.} \bibinfo{year}{2014}\natexlab{}.
\newblock \showarticletitle{Glove: Global vectors for word representation}. In
  \bibinfo{booktitle}{\emph{Proceedings of the 2014 conference on empirical
  methods in natural language processing (EMNLP)}}.
  \bibinfo{pages}{1532--1543}.
\newblock


\bibitem[\protect\citeauthoryear{Rousseau, Kiagias, and Vazirgiannis}{Rousseau
  et~al\mbox{.}}{2015}]%
        {rousseau2015text}
\bibfield{author}{\bibinfo{person}{Fran{\c{c}}ois Rousseau},
  \bibinfo{person}{Emmanouil Kiagias}, {and} \bibinfo{person}{Michalis
  Vazirgiannis}.} \bibinfo{year}{2015}\natexlab{}.
\newblock \showarticletitle{Text categorization as a graph classification
  problem}. In \bibinfo{booktitle}{\emph{Proceedings of the 53rd Annual Meeting
  of the Association for Computational Linguistics and the 7th International
  Joint Conference on Natural Language Processing (Volume 1: Long Papers)}},
  Vol.~\bibinfo{volume}{1}. \bibinfo{pages}{1702--1712}.
\newblock


\bibitem[\protect\citeauthoryear{Rousseau and Vazirgiannis}{Rousseau and
  Vazirgiannis}{2013}]%
        {rousseau2013graph}
\bibfield{author}{\bibinfo{person}{Fran{\c{c}}ois Rousseau} {and}
  \bibinfo{person}{Michalis Vazirgiannis}.} \bibinfo{year}{2013}\natexlab{}.
\newblock \showarticletitle{Graph-of-word and TW-IDF: new approach to ad hoc
  IR}. In \bibinfo{booktitle}{\emph{Proceedings of the 22nd ACM international
  conference on Information \& Knowledge Management}}. ACM,
  \bibinfo{pages}{59--68}.
\newblock


\bibitem[\protect\citeauthoryear{Rousseau and Vazirgiannis}{Rousseau and
  Vazirgiannis}{2015}]%
        {rousseau2015main}
\bibfield{author}{\bibinfo{person}{Fran{\c{c}}ois Rousseau} {and}
  \bibinfo{person}{Michalis Vazirgiannis}.} \bibinfo{year}{2015}\natexlab{}.
\newblock \showarticletitle{Main core retention on graph-of-words for
  single-document keyword extraction}. In \bibinfo{booktitle}{\emph{European
  Conference on Information Retrieval}}. Springer, \bibinfo{pages}{382--393}.
\newblock


\bibitem[\protect\citeauthoryear{Schlichtkrull, Kipf, Bloem, Berg, Titov, and
  Welling}{Schlichtkrull et~al\mbox{.}}{2017}]%
        {schlichtkrull2017modeling}
\bibfield{author}{\bibinfo{person}{Michael Schlichtkrull},
  \bibinfo{person}{Thomas~N Kipf}, \bibinfo{person}{Peter Bloem},
  \bibinfo{person}{Rianne van~den Berg}, \bibinfo{person}{Ivan Titov}, {and}
  \bibinfo{person}{Max Welling}.} \bibinfo{year}{2017}\natexlab{}.
\newblock \showarticletitle{Modeling Relational Data with Graph Convolutional
  Networks}.
\newblock \bibinfo{journal}{\emph{arXiv preprint arXiv:1703.06103}}
  (\bibinfo{year}{2017}).
\newblock


\bibitem[\protect\citeauthoryear{Simonyan and Zisserman}{Simonyan and
  Zisserman}{2015}]%
        {simonyan2015very}
\bibfield{author}{\bibinfo{person}{Karen Simonyan} {and}
  \bibinfo{person}{Andrew Zisserman}.} \bibinfo{year}{2015}\natexlab{}.
\newblock \showarticletitle{Very deep convolutional networks for large-scale
  image recognition}.
\newblock \bibinfo{journal}{\emph{Proceedings of the International Conference
  on Learning Representations}} (\bibinfo{year}{2015}).
\newblock


\bibitem[\protect\citeauthoryear{Srivastava, Hinton, Krizhevsky, Sutskever, and
  Salakhutdinov}{Srivastava et~al\mbox{.}}{2014}]%
        {srivastava2014dropout}
\bibfield{author}{\bibinfo{person}{Nitish Srivastava},
  \bibinfo{person}{Geoffrey Hinton}, \bibinfo{person}{Alex Krizhevsky},
  \bibinfo{person}{Ilya Sutskever}, {and} \bibinfo{person}{Ruslan
  Salakhutdinov}.} \bibinfo{year}{2014}\natexlab{}.
\newblock \showarticletitle{Dropout: A simple way to prevent neural networks
  from overfitting}.
\newblock \bibinfo{journal}{\emph{Journal of Machine Learning Research}}
  \bibinfo{volume}{15}, \bibinfo{number}{1} (\bibinfo{year}{2014}),
  \bibinfo{pages}{1929--1958}.
\newblock


\bibitem[\protect\citeauthoryear{Tixier, Malliaros, and Vazirgiannis}{Tixier
  et~al\mbox{.}}{2016}]%
        {tixier2016graph}
\bibfield{author}{\bibinfo{person}{Antoine Tixier}, \bibinfo{person}{Fragkiskos
  Malliaros}, {and} \bibinfo{person}{Michalis Vazirgiannis}.}
  \bibinfo{year}{2016}\natexlab{}.
\newblock \showarticletitle{A graph degeneracy-based approach to keyword
  extraction}. In \bibinfo{booktitle}{\emph{Proceedings of the 2016 Conference
  on Empirical Methods in Natural Language Processing}}.
  \bibinfo{pages}{1860--1870}.
\newblock


\bibitem[\protect\citeauthoryear{Vaswani, Shazeer, Parmar, Uszkoreit, Jones,
  Gomez, Kaiser, and Polosukhin}{Vaswani et~al\mbox{.}}{2017}]%
        {vaswani2017attention}
\bibfield{author}{\bibinfo{person}{Ashish Vaswani}, \bibinfo{person}{Noam
  Shazeer}, \bibinfo{person}{Niki Parmar}, \bibinfo{person}{Jakob Uszkoreit},
  \bibinfo{person}{Llion Jones}, \bibinfo{person}{Aidan~N Gomez},
  \bibinfo{person}{{\L}ukasz Kaiser}, {and} \bibinfo{person}{Illia
  Polosukhin}.} \bibinfo{year}{2017}\natexlab{}.
\newblock \showarticletitle{Attention is all you need}. In
  \bibinfo{booktitle}{\emph{Advances in Neural Information Processing
  Systems}}. \bibinfo{pages}{6000--6010}.
\newblock


\bibitem[\protect\citeauthoryear{Veli{\v{c}}kovi{\'c}, Cucurull, Casanova,
  Romero, Li{\`o}, and Bengio}{Veli{\v{c}}kovi{\'c} et~al\mbox{.}}{2017}]%
        {velivckovic2017graph}
\bibfield{author}{\bibinfo{person}{Petar Veli{\v{c}}kovi{\'c}},
  \bibinfo{person}{Guillem Cucurull}, \bibinfo{person}{Arantxa Casanova},
  \bibinfo{person}{Adriana Romero}, \bibinfo{person}{Pietro Li{\`o}}, {and}
  \bibinfo{person}{Yoshua Bengio}.} \bibinfo{year}{2017}\natexlab{}.
\newblock \showarticletitle{Graph Attention Networks}.
\newblock \bibinfo{journal}{\emph{arXiv preprint arXiv:1710.10903}}
  (\bibinfo{year}{2017}).
\newblock


\bibitem[\protect\citeauthoryear{Wang, Wu, Coates, and Ng}{Wang
  et~al\mbox{.}}{2012}]%
        {wang2012end}
\bibfield{author}{\bibinfo{person}{Tao Wang}, \bibinfo{person}{David~J Wu},
  \bibinfo{person}{Adam Coates}, {and} \bibinfo{person}{Andrew~Y Ng}.}
  \bibinfo{year}{2012}\natexlab{}.
\newblock \showarticletitle{End-to-end text recognition with convolutional
  neural networks}. In \bibinfo{booktitle}{\emph{Pattern Recognition (ICPR),
  2012 21st International Conference on}}. IEEE, \bibinfo{pages}{3304--3308}.
\newblock


\bibitem[\protect\citeauthoryear{Wu, Pan, Zhu, Zhang, and Philip}{Wu
  et~al\mbox{.}}{2017}]%
        {wu2017multiple}
\bibfield{author}{\bibinfo{person}{Jia Wu}, \bibinfo{person}{Shirui Pan},
  \bibinfo{person}{Xingquan Zhu}, \bibinfo{person}{Chengqi Zhang}, {and}
  \bibinfo{person}{S~Yu Philip}.} \bibinfo{year}{2017}\natexlab{}.
\newblock \showarticletitle{Multiple structure-view learning for graph
  classification}.
\newblock \bibinfo{journal}{\emph{IEEE transactions on neural networks and
  learning systems}} (\bibinfo{year}{2017}).
\newblock


\bibitem[\protect\citeauthoryear{Xu, Ba, Kiros, Cho, Courville, Salakhudinov,
  Zemel, and Bengio}{Xu et~al\mbox{.}}{2015}]%
        {xu2015show}
\bibfield{author}{\bibinfo{person}{Kelvin Xu}, \bibinfo{person}{Jimmy Ba},
  \bibinfo{person}{Ryan Kiros}, \bibinfo{person}{Kyunghyun Cho},
  \bibinfo{person}{Aaron Courville}, \bibinfo{person}{Ruslan Salakhudinov},
  \bibinfo{person}{Rich Zemel}, {and} \bibinfo{person}{Yoshua Bengio}.}
  \bibinfo{year}{2015}\natexlab{}.
\newblock \showarticletitle{Show, attend and tell: Neural image caption
  generation with visual attention}. In \bibinfo{booktitle}{\emph{International
  conference on machine learning}}. \bibinfo{pages}{2048--2057}.
\newblock


\bibitem[\protect\citeauthoryear{Yu and Koltun}{Yu and Koltun}{2016}]%
        {yu2015multi}
\bibfield{author}{\bibinfo{person}{Fisher Yu} {and} \bibinfo{person}{Vladlen
  Koltun}.} \bibinfo{year}{2016}\natexlab{}.
\newblock \showarticletitle{Multi-Scale Context Aggregation by Dilated
  Convolutions}. In \bibinfo{booktitle}{\emph{Proceedings of the International
  Conference on Learning Representations}}.
\newblock


\bibitem[\protect\citeauthoryear{Zeng, Liu, Lai, Zhou, and Zhao}{Zeng
  et~al\mbox{.}}{2014}]%
        {zeng2014relation}
\bibfield{author}{\bibinfo{person}{Daojian Zeng}, \bibinfo{person}{Kang Liu},
  \bibinfo{person}{Siwei Lai}, \bibinfo{person}{Guangyou Zhou}, {and}
  \bibinfo{person}{Jun Zhao}.} \bibinfo{year}{2014}\natexlab{}.
\newblock \showarticletitle{Relation classification via convolutional deep
  neural network}. In \bibinfo{booktitle}{\emph{Proceedings of COLING 2014, the
  25th International Conference on Computational Linguistics: Technical
  Papers}}. \bibinfo{pages}{2335--2344}.
\newblock


\bibitem[\protect\citeauthoryear{Zhang, Zhao, and LeCun}{Zhang
  et~al\mbox{.}}{2015}]%
        {zhang2015character}
\bibfield{author}{\bibinfo{person}{Xiang Zhang}, \bibinfo{person}{Junbo Zhao},
  {and} \bibinfo{person}{Yann LeCun}.} \bibinfo{year}{2015}\natexlab{}.
\newblock \showarticletitle{Character-level convolutional networks for text
  classification}. In \bibinfo{booktitle}{\emph{Advances in neural information
  processing systems}}. \bibinfo{pages}{649--657}.
\newblock


\end{thebibliography}

\end{document}